\documentclass[runningheads]{llncs}

\usepackage{eccv}

\usepackage{eccvabbrv}

\usepackage{graphicx}
\usepackage{booktabs}
\usepackage{amsfonts}       
\usepackage{nicefrac}       
\usepackage{microtype}      
\usepackage{xcolor}         
\usepackage{makecell}
\usepackage{balance}
\usepackage{caption}
\usepackage{diagbox}

\usepackage{amsmath}
\usepackage{amssymb}
\usepackage{multirow}
\usepackage{layouts}

\usepackage[T1]{fontenc}
\usepackage[utf8]{inputenc}
\usepackage{amsmath,amssymb}
\usepackage{xcolor}
\usepackage[english]{babel}

\usepackage{tabularx}
\newcolumntype{Y}{>{\centering\arraybackslash}r}

\usepackage[accsupp]{axessibility}  

\usepackage{hyperref}

\usepackage{orcidlink}

\begin{document}

\title{Adversarial purification for no-reference image-quality metrics: applicability study and new methods} 

\titlerunning{Adversarial purification for no-reference image-quality metrics}

\author{Gushchin Aleksandr\inst{1}\orcidlink{0000-0002-4055-7394} \and
Anna Chistyakova\inst{2}\orcidlink{1111-2222-3333-4444} \and
Vladislav Minashkin\inst{3}\orcidlink{1111-2222-3333-4444} \and
Anastasia Antsiferova\inst{1}\orcidlink{1111-2222-3333-4444} \and
Dmitriy Vatolin\inst{1}\orcidlink{2222--3333-4444-5555}}

\authorrunning{A.~Gushchin et al.}

\institute{Lomonosov Moscow State University
\email{\{alexander.gushchin,aantsiferova,dmitriy\}@graphics.cs.msu.ru}  \and
ISP RAS
\email{\{a.chistyakova\}@ispras.ru}  \and
Moscow Institute of Physics and Technology
\email{\{minashkin.vm\}@phystech.edu}}

\maketitle

\begin{abstract}
  Recently, the area of adversarial attacks on image quality metrics has begun to be explored, whereas the area of defences remains under-researched. In this study, we aim to cover that case and check the transferability of adversarial purification defences from image classifiers to IQA methods. In this paper, we apply several widespread attacks on IQA models and examine the success of the defences against them. The purification methodologies covered different preprocessing techniques, including geometrical transformations, compression, denoising, and modern neural network-based methods. Also, we address the challenge of assessing the efficacy of a defensive methodology by proposing ways to estimate output visual quality and the success of neutralizing attacks. We test defences against attacks on three IQA metrics -- Linearity\cite{linearity}, MetaIQA\cite{metaiqa} and SPAQ\cite{spaq}. The code for attacks and defences is available at: \textit{link is hidden for a blind review}.

  \keywords{Adversarial robustness \and Image quality assessment \and Adversarial purification}
\end{abstract}


\section{Introduction}
\label{sec:intro}

Learning-based image quality assessment (IQA) metrics are widely used to develop and evaluate image and video processing algorithms. Compared to traditional metrics based on error visibility or structural similarity paradigms \cite{duanmu2021quantifying}, learning-based metrics show a higher correlation with subjective quality. However, several studies showed that learning-based image-quality metrics are vulnerable to adversarial perturbations, which may lead to several negative impacts. For example, IQA metrics are usually published as a part of open benchmarks evaluation methodology for reproducibility. The participants may construct adversarial attacks on metrics to receive higher positions on the leaderboard. Thus, IQA adversarial robustness has started to develop. This area is not as well-studied as the robustness of image classification or detection methods. For image quality metrics, only empirical experiments with adversarial attacks have been conducted so far and no provable results have been received. Among existing adversarial attacks proposed to evaluate metrics' robustness, there are several methods based on gradient optimisation \cite{zhang2022perceptual, shumitskaya2022universal, shumitskaya2024towards}, perceptual-oriented masking \cite{Korhonen2022AdversarialAA, Luo_2022_CVPR} and image enhancement \cite{zvezdakova2019hacking}. 

The defence mechanisms for image—and video-quality metrics are almost not studied yet. There are two reasons for this: first, the area of attacks on IQA metrics is in the early stage of its development, and second, a problem formulation for defending IQA metrics is more challenging than, for example, image classification. Image semantics stays invariant after adding an adversarial perturbation to an image, so the true label of an adversarial image is still known. However, its perceptual quality changes and the defence method needs to make the original quality be estimated correctly. 
To our knowledge, only two metrics were positioned as more robust versions of the original implementations. The first was VMAF NEG \cite{li2021vmaf_neg} by Netflix, which increases VMAF's stability to image enhancement by clipping scores of base SVM features. R-LPIPS \cite{ghazanfari2023r} and E-LPIPS \cite{kettunen2019lpips} are robust versions of the LPIPS metric created using adversarial training and ensembling. 

\begin{figure*}[tb]
\centering
  \centering
   \includegraphics[width=0.7\linewidth]{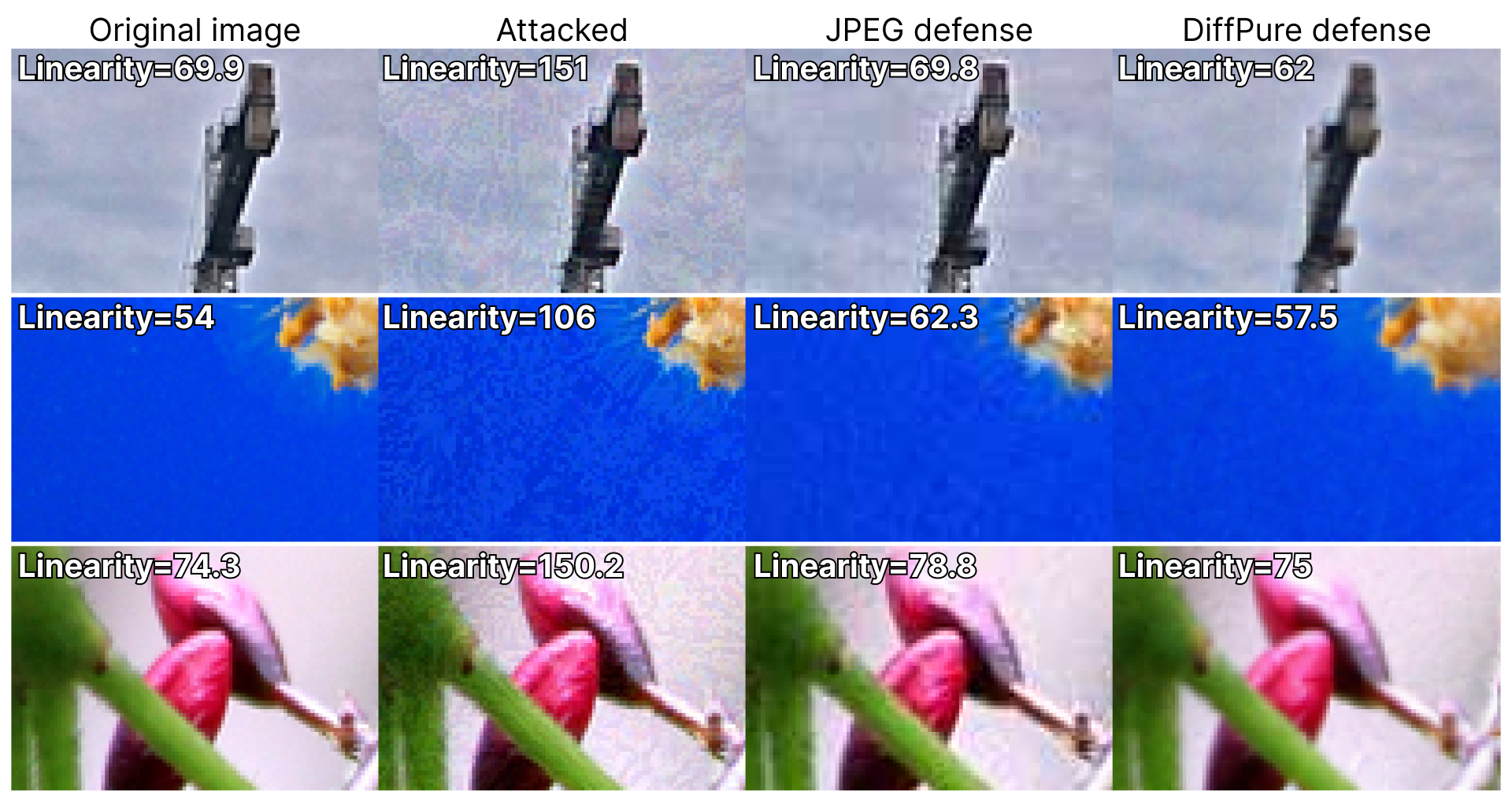}
  \caption{Original image (first column), image after the adversarial attack (AMI-FGSM, second columns), and two defence techniques applied to the adversarial image (third and fourth columns). All images are cropped to size $100\times80$.}
  \label{fig:defence_example}
\end{figure*}

Our study focuses on investigating and improving the robustness of no-reference metrics for several reasons. First, these metrics are easier to attack; they have more practical applications due to the optionality of accessing the original image (e.g. image and video generation, video streaming, etc.). No-reference metrics can be integrated into the loss function in various computer vision tasks. Finally, the above examples of robust metrics are full-reference ones; no robust versions were proposed for no-reference (NR) metrics, and only a few experiments were conducted. 
Korhonen \& You \cite{Korhonen2022AdversarialAA} tried basic defence methods for IQA models (detection, adversarial training and preprocessing) and showed that adversarial training helps to improve the empirical robustness. Image cropping and resizing effectively against universal adversarial perturbations in \cite{shumitskaya2024towards}. Figure \ref{fig:defence_example} shows an example of attacked IQA metric Linearity and different preprocessing defences against this attack.

Robust metrics are essential for developing contemporary image processing and compression methods. Such metrics will lead to the development of trusted benchmarks and allow researchers to use metrics as an optimization component to train processing methods and reduce costly subjective tests. The most popular defence methods for neural networks are adversarial training, adversarial purification, and attack detection. This paper focuses on empirical defence mechanisms as a first step towards improving metrics' robustness. Despite adversarial training being efficient, it also proved to reduce metrics' performance. This work studies the applicability of image purifications to defending IQA metrics and proposes new purification methods. Our contributions are as follows. First, we propose a new dataset of adversarial images for three IQA metrics by 10 attacks. Second, we propose two new purification methods: one based on the combination of DiffPure \cite{nie2022diffusion} and unsharp masking and edging techniques that show state-of-the-art performance on restricted adversarial attacks, and another called FCN filter that is efficient against unbounded colour filtering attack AdvCF \cite{advcf}. Finally, we provide the results and analysis of extensive experiments involving 16 purification methods against 10 attacks.

\section{Related work}

\subsection{Adversarial attacks on IQA}
Szegedy et al. \cite{Szegedy2013IntriguingPO} showed that neural networks are vulnerable to specially generated input data, leading to incorrect model output. This later became known as an \textit{adversarial attack}. A large number of attack methods have been proposed, particularly for the task of image classification. Most of these attacks \cite{goodfellow2015explaining, kurakin2017adversarial, madry2019deep, carlini2017evaluating} are additive, as additional adversarial noise is generated and added to the original image. These methods solve an optimization problem in pixel space, with some restrictions on perturbation magnitude in terms of the $l_p$ norm. Since $l_p$ was shown to be inefficient in approximating visual quality, perceptual-quality-preserving adversarial attacks and colour transformation techniques were proposed \cite{advcf}.

Despite the majority of existing adversarial attacks being created for computer vision models, attacks are developed specifically for IQA metrics. AMI-FGSM \cite{amifgsm} is an attack on NR metrics that extends FGSM \cite{goodfellow2015explaining}. Shumitskaya et al. \cite{shumitskaya2022universal} proposed training a universal perturbation on a set of images. Ghildyal \& Liu \cite{ghildyal2023attacking} highlighted the vulnerability of full-reference (FR) metrics to adversarial attacks. They adapted common adversarial attacks such as FGSM, PGD, and the one-pixel attack to generate adversarial examples on IQA metrics. Korhonen \& You \cite{Korhonen2022AdversarialAA} proposed an iterative attack on no-reference metrics that improves the visual quality of attacked images using the Sobel filter.

\subsection{Adversarial defences for neural networks}
\label{RelWorkdefences}

Similar to attack methods, many defences have been proposed for the image classification task. Modifying the training or inference process can improve robustness to adversarial attacks. One of the most popular methods of defending models against adversarial attacks is adversarial training \cite{goodfellow2015explaining, madry2019deep}. It involves the dynamic addition of adversarially generated examples to the training set. This approach not only requires a modification of the training procedure but also increases the computational cost, as at each iteration, it is necessary to perform an additional one or even several forward and backward passes to generate an adversarial example. Besides, adversarial training leads to a performance decrease on clean inputs. A similar effect of adversarial training was shown by Korhonen \& You \cite{Korhonen2022AdversarialAA} for IQA models.

Another defence type is preprocessing input data with auxiliary transformations before the primary model, also called purification. This type of defence necessitates modifying the model's inference.
Graece et al. \cite{graese2016assessing} proposed a method for adversarial purification based on simple image transformations such as cropping and blurring. Guo et al. \cite{JPEG_defence3} suggested defending against adversarial attacks by averaging the predictions of multiple random image crops during evaluation. This approach aims to alter the spatial location of the adversarial perturbations, which are not invariant to such transformations. Several works \cite{JPEG_defence1, JPEG_defence2, JPEG_defence3} suggested JPEG compression as a method for preventing adversarial attacks.
Xu et al. \cite{DBLP:journals/corr/XuEQ17} proposed a feature-squeezing approach for reducing the dimensions of the input image to decrease the attacker's capability to conduct an adversarial attack. The main idea is to "squeeze out" unnecessary features by reducing the colour bit depth or applying spatial smoothing. Guo et al. \cite{JPEG_defence3} studied pixel dropout with total variation minimization \cite{Rudin1992NonlinearTV}. This approach involves selecting a small set of pixels and reconstructing the image without adversarial noise. 
Meng \& Chen, in their MagNet defence \cite{magnet}, used an auto-encoder to move adversarial examples closer to the manifold of clean examples. However, this method is intended for scenarios where the attacker cannot access the model parameters and, therefore, is unsuitable for gradient-based attacks, which constitute a significant threat.
As the popularity of generative models has grown, they have also been used for adversarial purification. In particular, Samangouei et al. \cite{defencegan} proposed using generative adversarial networks (GANs) that remove adversarial perturbations from images. Recently, Nie et al. \cite{nie2022diffusion} proposed using increasingly popular diffusion models to remove adversarial perturbations from images. The idea is that the adversarial perturbations should blend with the noise in the forward diffusion process, and then the reverse process removes both the noise and the perturbations, resulting in a pure image.

Not all purification methods for classifiers are suitable for metric models, as image quality is not a decisive factor in classifier performance. In the IQA task, it is important not only to return the original metric scores but also not to alter the image significantly. Only a few defence methods have been proposed for these models. Korhonen et al. \cite{Korhonen2022AdversarialAA} employed Gaussian blurring and bilateral filtering against adversarial perturbations. The authors conclude that preprocessing methods are worthy of detailed study in the future, which is presented in our work. 

\section{Methodology}
\subsection{Problem definition}
Our work considers attacks that aim to increase predicted quality scores. These attacks have been suggested in recent papers and have greater practical applicability \cite{Korhonen2022AdversarialAA, shumitskaya2022universal}. The attacks that aim to reduce the predicted image quality look the same, except for the vector sign that represents the attack's direction. Thus, this approach does not violate the generality of the study.

For a given no-reference image quality metric $f$ and an input image $x \in \mathbb{R}^{H \times W \times 3}$, an adversarial attack that aims to increase the predicted quality score can be mathematically described as follows:
\begin{equation}
    \max f(att(x)), \;\text{such that}\; dist(x, att(x)) \leq \varepsilon,
\end{equation}
where $att: \mathbb{R}^{H \times W \times 3} \rightarrow \mathbb{R}^{H \times W \times 3}$ is the adversarial attack algorithm applied to $x$, $dist(\cdot)$ is some distance metric and $\varepsilon$ is a constraint on the distance. An attack is successful if it increases the predicted metric score but does not improve the visual quality. Therefore, the changes in the image should be invisible.

Adversarial defence is a process to reduce the quality score improvement caused by an attack: 
\begin{equation}
    \min \Bigl(|f(g(att(x))) - f(x)| + \lambda dist(x, g(att(x)))\Bigr)
\end{equation}
where $g: \mathbb{R}^{H \times W \times 3} \rightarrow \mathbb{R}^{H \times W \times 3}$ is a defence method. The adversarial purification defence aims to reverse the adversarial image to the original without reducing the correlation with subjective quality.

\subsection{Adversarial attacks}
\label{sec:list_of_attacks}

We selected 10 attack methods to evaluate adversarial purification defences. Most of them were originally introduced as attacks on image classification models; thus, we adapted them to the IQA task by replacing the loss function that increases metric score: \begin{equation}\label{attack_loss}\mathcal{L}(\theta, x) = 1 - \frac{f_\theta(x)}{max(f_\theta) - min(f_\theta)},\end{equation} where $max(f_\theta)$ and $min(f_\theta)$ represent the highest and the lowest objective scores respectively obtained by metric on the NIPS 2017: Adversarial Learning Development Set \cite{nips-dataset}. 

We describe the attacks we use in our work below. The first group of attacks is the Fast Sign Gradient Method (FGSM) and its variations. \textbf{FGSM} \cite{goodfellow2015explaining} takes one step in the opposite direction of the gradient to minimize the objective function, which in our case is \ref{attack_loss}. Its modification \textbf{I-FGSM} \cite{kurakin2017adversarial} creates adversarial examples, making several iterations of small steps toward the gradient. \textbf{MI-FGSM} \cite{dong2018boosting} additionally uses a momentum term during optimisation. \textbf{AMI-FGSM} \cite{Sang2023} was introduced as an attack on IQA models and represents a version of MI-FGSM that adjusts the allowable attack magnitude by limiting the perceptual quality computed with the no-reference metric NIQE \cite{Mittal2013MakingA}. 

Other selected methods employ various techniques to ensure the attack remains invisible to the human eye. \textbf{Korhonen et al.} \cite{Korhonen2022AdversarialAA} is a method for generating adversarial images for NR quality metrics using a spatial activity map to concentrate perturbations in textured regions. \textbf{SSAH} \cite{Luo_2022_CVPR} locates perturbations within high-frequency components using low-frequency constraints. \textbf{MADC} \cite{Wang} is a method for comparing image quality metrics. It synthesizes a pair of images to maximize or minimize the output of one IQA metric while keeping the other metric score constant. We use MADC as an attack on IQA metric by moving towards the direction of the metric at fixed MSE values, as it was proposed in Antsiferova et al. \cite{antsiferova2024comparing}. \textbf{Zhang et al.} \cite{zhang2022perceptual} suggested adding a full-reference metric as an additional term of the objective function to create an attack on NR IQA metrics. We employed three versions of this attack, where SSIM \cite{ssim}, LPIPS \cite{zhang2018unreasonable}, and DISTS \cite{ding2020image} metrics were used to preserve visual quality. 

The attacks described above are restricted because perturbations added to the adversarial image are limited by the $l_p$ norm or other quality metrics. We additionally use an unrestricted colour filter attack \textbf{AdvCF} \cite{advcf}, which changes colours in an image by optimizing a simple colour filter commonly used in popular photo editing software.

\subsection{Purification methods}
We investigate several standard preprocessing techniques for classifiers to purify images from additive adversarial noise. The first method we chose is the \textbf{resize} technique, which changes the resolution of input images to a smaller one and then resizes them back to their original resolution. We conducted experiments with different interpolation modes. The bilinear mode was chosen as the baseline. Following Guo et al. \cite{JPEG_defence3}, we randomly \textbf{cropped} and resized images to their original size. We also considered \textbf{JPEG} defence that is widely used for image classification defence. Also, our list of methods includes spatial transformations such as \textbf{flipping} (mirroring the image) and \textbf{random rotation}. Similar to the experiments for IQA performed by Korhonen \& You. \cite{Korhonen2022AdversarialAA}, we applied Gaussian \textbf{blurring} and \textbf{bilateral filtering}. The Gaussian filter smoothes the entire image without considering details. In contrast, the bilateral filter, which is non-linear, accounts for the intensity values of neighbouring pixels, thus preserving edges during smoothing. In addition, we evaluated the \textbf{median filter}, which replaces each pixel with the median of its neighbouring pixels. Also, we added a composition of \textbf{Unsharp} masking and Gaussian blurring techniques. The intuition behind this method is as follows: first, we denoise the images and filter out high-frequency noise, and then we sharpen the edges of objects. Thus, the first phase is for filtering the adversarial perturbation, and the second phase is for restoring the quality of the purified image.
\textbf{DiffPure} \cite{nie2022diffusion} diffusion model shown to be effective in purifying adversarial images. A small amount of noise is introduced to the adversarial example through forward diffusion. The diffusion process continues until an optimally computed time step is reached. Subsequently, the model reverses this diffusion to recover a purified, clean image. The goal is for the perturbations to gradually merge with the noise, allowing the added Gaussian noise to dominate. As a result, the reverse process removes the added noise and the adversarial perturbations, resulting in a purified image.

Due to the blurring effect created by DiffPure and considering the penalties imposed by quality metrics for blurring, we implemented two additional variants of add-ons to address this issue. The first one is called \textbf{DiffPure+Edge} and is based on blending adversarial and purified images. We detect edges within the image via the Sobel filter, then replace edge pixels with edges from the attacked image. This helps to sharpen the image without increasing adversarial perturbation. The second (\textbf{DiffPure+Unsharp}) is similar to the Unsharp method. First, DiffPure is applied to the adversarial image to remove all undesired noise, and then unsharp masking sharpens the image, improving its visual quality.

Additionally, we explore popular image restoration methods, such as the Multi-Stage Progressive Image Restoration Network (\textbf{MPRNet}) \cite{Zamir2021MPRNet} and \textbf{Real-ESRGAN} \cite{wang2021realesrgan}, for adversarial image purification. MPRNet \cite{Zamir2021MPRNet} is a three-stage convolutional neural network designed to solve three image restoration tasks such as deblurring, deraining, and denoising. The first two stages of the network use an encoder-decoder architecture to extract multi-scale contextual information. In contrast, the final stage operates on the image at its original resolution to preserve fine details. Key features of MPRNet include supervised attention modules placed between stages and cross-stage feature fusion, which ensure effective information transfer from the early to late stages. Real-ESRGAN \cite{wang2021realesrgan} is a restoration model based on a generative adversarial network trained with synthetic data. The authors used a deep network consisting of several residual dense blocks to perform super-resolution. We set a scaling factor of 1 to keep the original image size.

Previously, all defences were designed to reduce images' adversarial noise, making them ineffective against colour attacks, such as AdvCF \cite{advcf}. We propose a new defence method based on a compact, fully convolutional neural network (\textbf{FCN filter}). Our model consists of three convolutional layers that apply 64, 16, and 3 filters and preserve the original image dimensions. The dataset is 200 images from NIPS 2017 attacked by AdvCf, of which the train part is 80\%, and the validation part is the remaining 20\%. We train it using the Adam optimizer for 200 epochs with a learning rate 1e-3 to recover the original image from an attacked image, optimizing the Mean Squared Error (MSE). We save the best-performed model on the validation set regarding the SSIM score between purified and original images.

\subsection{Adversarial dataset}

We created a dataset with adversarial images to analyse the efficiency of defences against analysed attacks. Images from NIPS 2017: Adversarial Learning Development Set \cite{nips-dataset} were used as reference images that were further perturbed. This dataset was previously used for a wide range of computer vision tasks, including attacks on IQA metrics \cite{robustness-benchmark}. Linearity was attacked on each image with each attacked method from \cref{sec:list_of_attacks} using default parameters. Thus, from each clear image, we received 10 perturbed images. The examples of perturbed images are presented in Fig.~\ref{fig:attack_examples}. The table of all parameters that were used in the attacks is presented in Supplementary material.

\begin{figure*}
\centering
  \centering
   \includegraphics[width=0.8\linewidth]{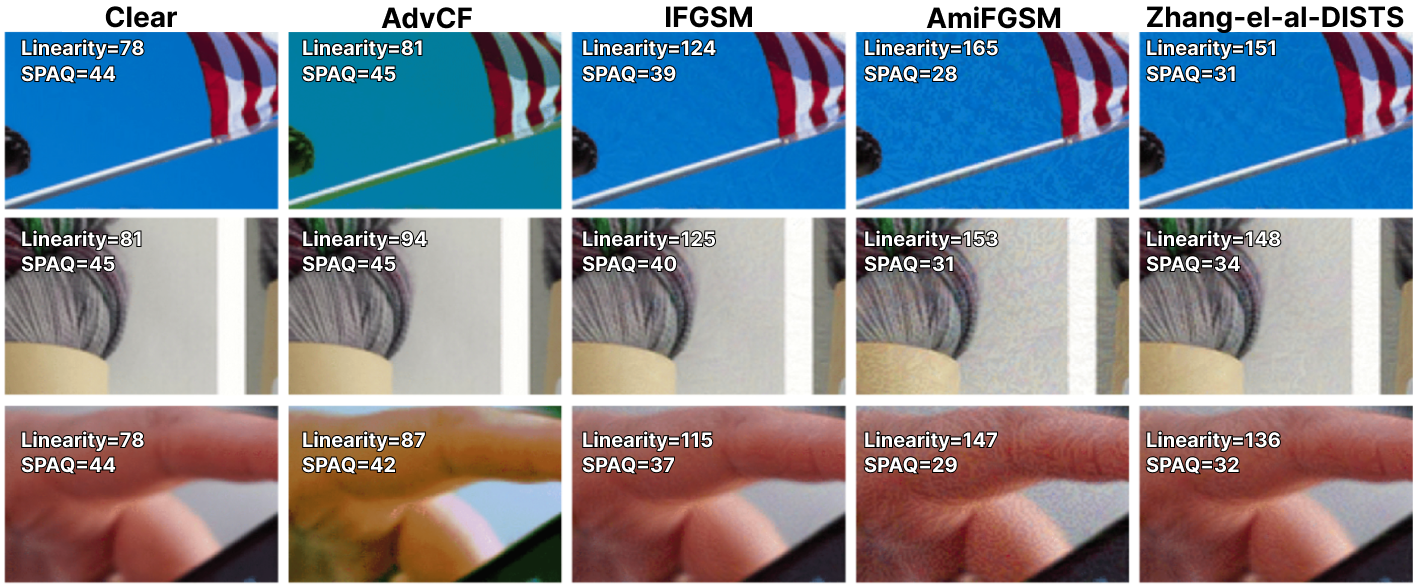}
  \caption{Examples of attacks. The first column shows the original (clean) image, the following columns display the corresponding image after the attack.}
  \label{fig:attack_examples}
\end{figure*}

As a target of adversarial attacks, we used three no-reference image quality metrics --  Linearity \cite{linearity}, MetaIQA \cite{metaiqa} and SPAQ \cite{spaq}. Results in this paper presented only for Linearity metric, results for MetaIQA and SPAQ can be found in SUpplementary material. The authors of Linearity created their own loss function, called ``norm-in-norm'', which converges about 10 times faster than the popular MAE and MSE loss functions. Linearity was chosen as the target metric because it shows high performance (correlation with subjective quality and speed) and medium robustness to adversarial attacks:

\begin{itemize}
    \item Correlation with subjective scores and high speed. In MSU No-Reference Video Quality Metrics Benchmark \cite{antsiferova2022video} Linearity shows the best performance among all no-reference image quality metrics. Its SROCC with subjective quality score is 0.91. Linearity also has a high calculation speed compared to its counterparts \cite{metrics-benchmark}. 
    \item Medium robustness. For our study, we needed a target IQA metric of medium robustness to adversarial attacks. Attacking highly robust metrics will lead to adversarial images that differ greatly from clear ones, or the attack will rather have poor success in increasing metrics' scores. Thus, the practical utility of these attacks is negligible. When targeting metrics of low robustness, the resulting attacks will be visually imperceptible. Such attacks can be neutralized by applying Gaussian blur with a small kernel size. According to \cite{robustness-benchmark}, Linearity shows medium robustness compared to other metrics, making it a valid target for our work.
\end{itemize}

Supplementary materials present an analysis of adversarial images. We calculate several characteristics and compare clear and perturbed images. 

\subsection{Implementation details}
We used public source code for Linearity without additional pre-training and selected the default parameters. The list of attack hyper-parameters and their choice justification is in the Supplementary materials. To ensure complete reproducibility of the results, we implemented a sophisticated end-to-end pipeline that handles the entire computation from metrics scores, attacks, and defence applications to performance metrics evaluation. 
Calculations were made on a computer with the following characteristics: NVIDIA A100-PCIE-40GB, an Intel Xeon Processor (Icelake) 32-core Processor @ 2.90GHz. All calculations took a total of about 50 GPU hours.

\subsection{Evaluation metrics} 
We use two groups of metrics to evaluate the defence success and the perceptual quality of a purified image to compare the defence methods.

\textbf {Quality metrics} estimate the effectiveness of the defence in processing the image with respect to visual quality. We chose PSNR and SSIM image quality metrics due to their greater robustness to adversarial attacks than No-Reference metrics. Additionally, the non-differentiability of these metrics provides a more challenging task for attackers. To merge these two metrics into one score, we simply add them with linear coefficient:
\begin{equation}
Quality ~ score = \frac{PSNR}{100} + \frac{SSIM}{2}
\end{equation}
These coefficients were chosen for the following reasons: for SSIM, the maximum value is 1. PSNR with values greater than 50 means that the compared images are completely indistinguishable, although PSNR has no maximum value. 

\textbf {Score metrics} reflect the effectiveness of the defence regarding the metric scores.
We calculate the relative gain for attacked and purified (after consecutive attack and defence) images as follows: 
\begin{equation}
Gain ~ score = \frac{|Clear - Distorted|}{Clear},
\end{equation} where $Distorted$ is either an attacked or purified image corresponding to a clear image.
We also calculate the Spearman correlation coefficient (SROCC) between metric scores on clean and purified images.

\section{Results}
In this section, we present results for defences against the Linearity metric. Results MetaIQA and SPAQ can be found in the Supplementary material.
We distinguish three main characteristics that defences should possess: neutralizing the effects of adversarial attacks, preserving the original image quality, and maintaining correlations of the values of the attacked model. To evaluate these features, we compare purification methods by Relative gain, PSNR/SSIM, and Spearman correlation coefficient (SROCC) ou the adversarial dataset described in Section 3.4.

\textbf{Overall defences efficiency}. \cref{tab:main_all} shows the results for clear and attacked images combined. Upscale, MPRNet, and Unsharp are the leaders in terms of output image quality. Nevertheless, DiffPure and DiffPure+Unsharp show comparable performance, falling less than 4\% behind the leader in terms of both PSNR and SSIM. The following methods are best at neutralizing the effects of the attack -- DiffPure+Unsharp, DiffPure, and resize. Also, DiffPure and DiffPure+Unsharp, alongside bilateral filter, maintain the highest correlations between metric values. 
\begin{table}[htb]
   \centering
   \caption{Results of adversarial purification defences averaged across clear and adversarial images.}
   {
   \small
   
    \begin{tabular}{@{}lrrrrrr@{}}
        \toprule
        Defence & \makecell{Quality \\ score}$\uparrow$ & \makecell{Gain \\ score}$\downarrow$ & \makecell{PSNR}$\uparrow$ & \makecell{SSIM}$\uparrow$ & \makecell{SROCC \\ purified}$\uparrow$ & FPS$\uparrow$ \\
        \midrule
        Bilateral & 0.594 & 0.151 & 28.310 & 0.835 & \textbf{0.731} & 6.78\\
        Flip & 0.158 & \textbf{0.077} & 12.402 & 0.162 & 0.728 & \textbf{36.67}\\
        Blur & 0.625 & 0.138 & 29.036 & 0.887 & 0.471 & 16.65\\
        JPEG & 0.626 & 0.099 & 31.382 & 0.859 & 0.725 & 15.76\\
        Median filter & 0.615 & 0.118 & 30.617 & 0.848 & 0.671 & 16.56\\
        MPRNet & \textbf{0.653} & 0.179 & \textbf{32.484} & \textbf{0.900} & 0.611 & 6.33 \\
        Crop & 0.227 & 0.247 & 13.532 & 0.285 & 0.626 & 29.92 \\
        Real-ESRGAN & 0.636 & 0.360 & 29.876 & \textbf{0.899} & 0.562 & 10.15 \\
        Resize & 0.626 & 0.099 & 31.382 & 0.859 & 0.725 & \textbf{63.31}\\
        Rotate & 0.344 & 0.239 & 20.380 & 0.434 & 0.501 & \textbf{36.18} \\
        Unsharp & \textbf{0.649} & 0.133 & \textbf{32.195} & \textbf{0.896} & 0.513 & 15.96 \\
        Upscale & 0.632 & 0.207 & 29.586 & 0.894 & 0.586 & 35.50 \\
        \midrule
        FCN filter (proposed) & 0.591 & 0.409 & 26.387 & 0.852 & 0.619 & 35.56\\
        \midrule
        DiffPure & 0.628 & \textbf{0.093} & 31.219 & 0.866 & \textbf{0.746} & 3.08\\
        +edge (proposed) & \textbf{0.640} & 0.195 & \textbf{32.220} & 0.878 & 0.480 & 2.98\\
        +Unsharp (proposed) & 0.632 & \textbf{0.082} & 31.557 & 0.870 & \textbf{0.750} & 2.90\\
        \bottomrule
        \end{tabular}
    }
   \label{tab:main_all}
 \end{table}

A purification that consists only of flipping the image along both axes shows the best Gain score. So, to obtain the metric value before the attack, one could simply calculate it on the flipped image. However, it is easy to optimise any attack against flipping defence.
Real-ESRGAN shows relatively good output quality but fails to mitigate the effects of adversarial attacks and to preserve the stability of metric values in terms of SROCC. The method works well in low-frequency areas but creates visually disturbing artefacts in high-texture regions.
DiffPure+Edge got the top-quality result but revealed a notably low SROCC between the obtained metric values and the Gain score. The reason behind this may be the strong emphasis on edges when calculating the values of the Linearity metric. 

Two outsiders in all three categories are random crop and rotate. We calculate SSIM and PSNR to evaluate output quality, relying heavily on pixel correspondence, as these metrics are full-reference. Because of that, it is obvious why crop and rotate show the worst quality score. Future research could explore how to properly assess the quality of such methods without relying on full-reference metrics.

To sum up, DiffPure seems like the preferred method to use in terms of all three compared characteristics. Also, adding unsharp masking after DiffPure slightly increases all of them. 
The primary drawback of this technique resides is its high computational complexity, leading to extended runtime. \cref{tab:main_all} also shows the average computational speed for each purification technique.

\textbf{Defences applied to clear images.} Because the defence does not know if an input image is perturbed or not, we check the Linearity performance after all defences on clean images. Regarding metric values stability, simple transformations such as rotation and cropping show the best results when applied to clean images. FCN filter is not far behind in the third place. They demonstrate remarkably good SROCC up to 0.97 while not significantly modifying the absolute values of the metric with a Gain score below 0.04. Regarding quality, however, rotation and cropping fall at the bottom of the rating, while the FCN filter has a high SSIM average value above 0.9. The leaders are four methods with very similar results -- DiffPure+Edge, MPRNet, unsharing and upscaling. A table with more detailed data on this case is presented in the Supplementary material.

\begin{figure*}
\centering
  \centering
   \includegraphics[width=\linewidth]{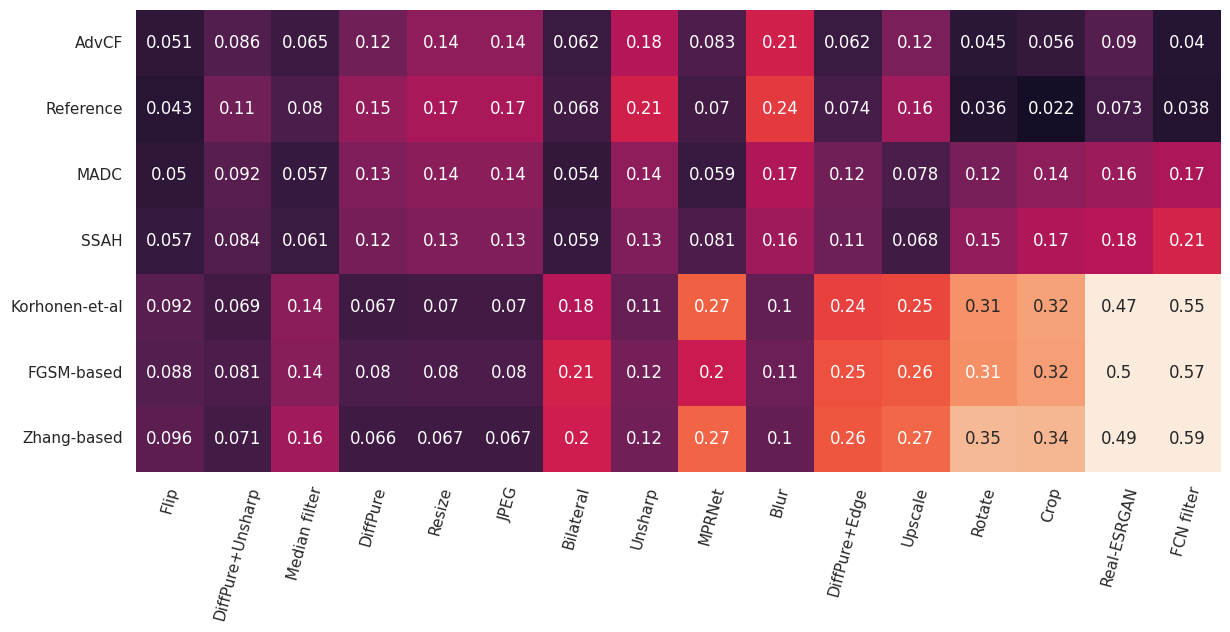}
  \caption{Heatmap showing Gain score for all attacks separately and for the original images. Purification methods are located along the X-axis. The Y-axis contains different attacks. Columns are sorted by their mean value. Thus, the more left the defence is, the better it eliminates the effects of the attack on average across all images. }
  \label{fig:gain_heatmap}
\end{figure*}

\textbf{Purification efficiency separately for each attack}. \cref{fig:gain_heatmap} illustrates the effectiveness of each defence in mitigating the impact of every attack, including a scenario with no attack. It can be seen that against weak attacks (AdvCF, MADC, SSAH), defences perform in a similar way with close Gain scores. For strong attacks, the gap between different defences is much bigger. Gaussian blur performs better on strong attacks but has worse Gain scores against weak attacks and clean images. It most likely corrupts the quality too much, and the metric value drops below the original value on a clean image. Balance for purification techniques can be achieved by varying their parameters.

\begin{figure*}
\centering
  \centering
   \includegraphics[width=\linewidth]{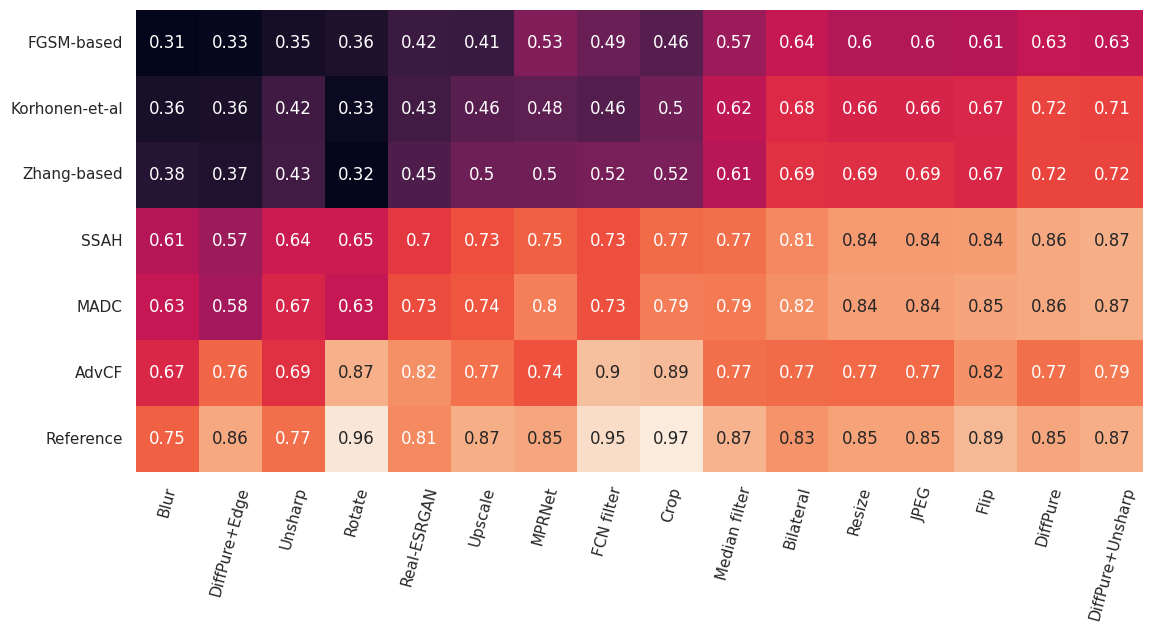}
  \caption{Heatmap showing Spearman correlation coefficients (SROCC) for all attacks separately and for the original images. Purification methods are located along the X-axis. The Y-axis contains different attacks. Columns are sorted by the mean value.}
  \label{fig:srocc_purified_heatmap}
\end{figure*}

A similar distinction between weak and strong attacks can be seen in \cref{fig:srocc_purified_heatmap}. The most stable defence regarding SROCC values turned out to be the bilateral filter. It shows SROCC above 0.68 for all types of attacks. Rotating has nearly perfectly maintained the stability of metric values on clean images, but completely falls behind on strong attacks. 

\begin{figure*}
\centering
  \centering
   \includegraphics[width=\linewidth]{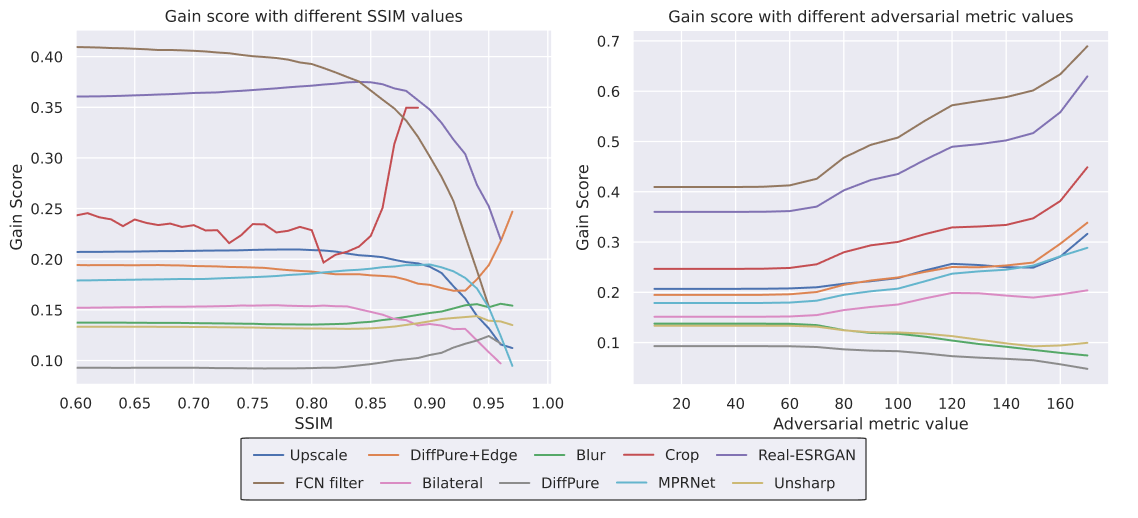}
  \caption{Gain score depending on minimum SSIM (left) / metric (right) values. Most defences modify their behaviour in terms of Gain score slightly depending on SSIM, but monotonically depends on adversarial metric values. Furthermore, most defences increase the Gain score with metric value, but some (DiffPure, Blur and Unsharp) start decreasing.}
  \label{fig:ssim_pretty}
\end{figure*}

\textbf{SSIM of purified images}. \cref{fig:ssim_pretty} (left) illustrates Gain score values for images with different minimal SSIM. The Gain score is equal to the average Gain score across the whole dataset. Some purification techniques were excluded for better visual representation. For most methods, the Gain score remains stable for all SSIM values. Some methods, such as Upscale, FCN filter and DiffPure+Edge behave differently against weak attacks. \cref{fig:ssim_pretty} (right) shows Gain score values based on minimal adversarial metric value. On the far right, the score was calculated only for strong attacks. On the far left, it was calculated on the whole dataset. We see that the Gain score changes monotonically when compared to the adversarial metric value. Three methods (DiffPure, Blur and Unsharp) stand out from the rest. Their Gain score monotonically decreases, showing that they are better at mitigating the effects of strong attacks.

\begin{figure*}
\begin{subfigure}{.54\textwidth}
  \centering
   \includegraphics[width=0.99\linewidth]{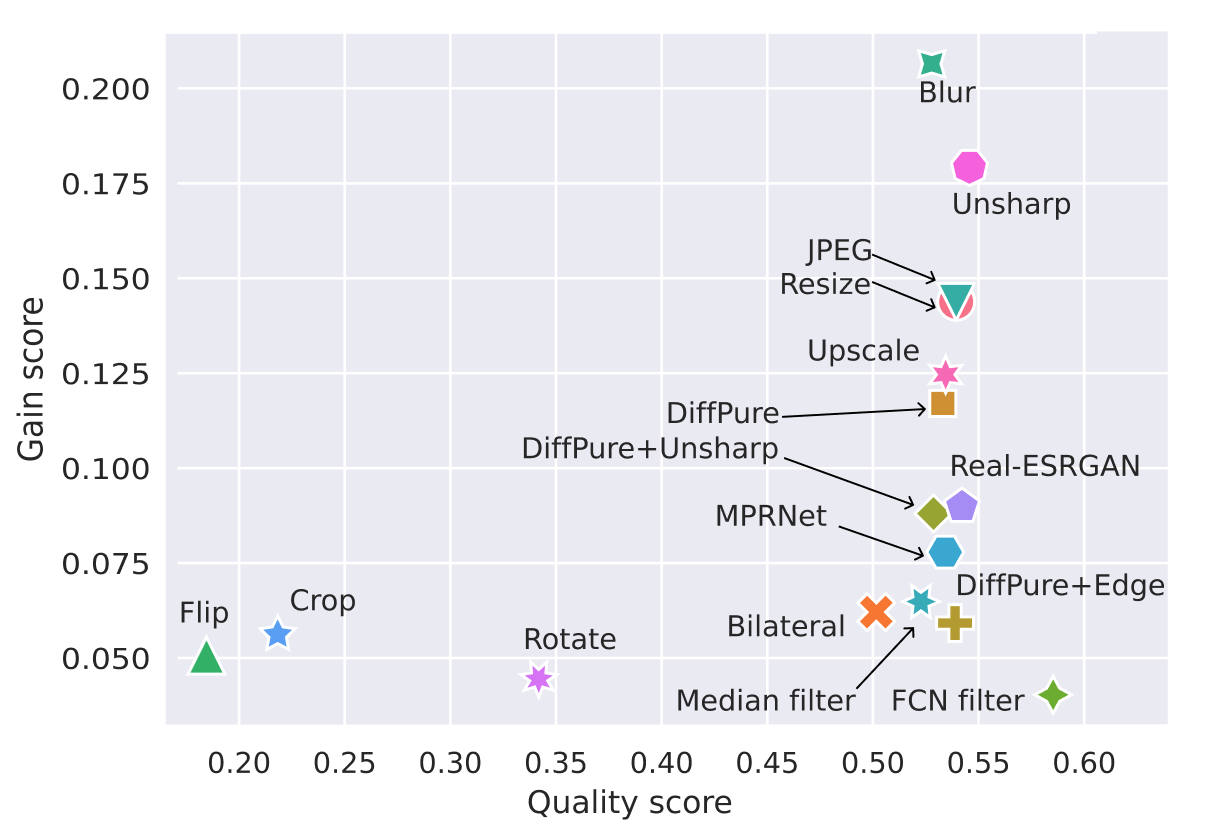}
  \caption{Quality/Gain scores for defences against \\ AdvCF attack.}
  \label{fig:advcf}
\end{subfigure}
  \hfill
\begin{subfigure}{.46\textwidth}
  \centering
   \includegraphics[width=0.99\linewidth]{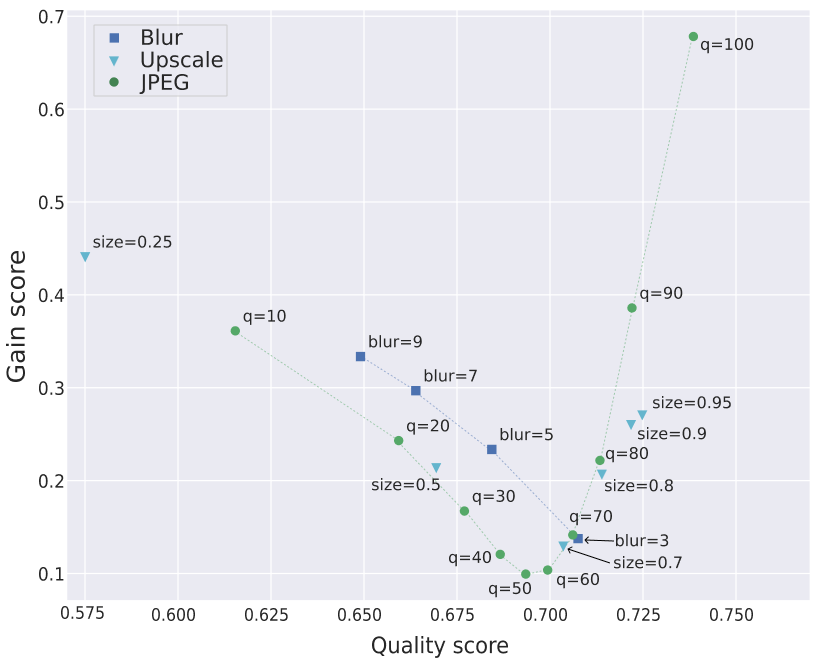}
  \caption{Gain and quality score for different parameters of blur, upscale and JPEG defences.}
  \label{fig:params}
\end{subfigure}
\end{figure*}

\textbf{Defence against unrestricted attack (AdvCF)}. All purification methods are more efficient against AdvCF on average than against all other attacks. \cref{fig:advcf} reveals that in this case, the overall leader in all categories is the proposed FCN filter. DiffPure-family methods also show stable performance across all categories and are not far behind the leader.

\textbf{The influence of parameters of blurring and upscaling}. \cref{fig:params} contains a comparison of three adversarial purification methods with different parameters. We computed JPEG compression with different quality factor from 10 to 100, gaussian blur with kernel size from 3 to 9 and upscale with size of upscaling image from 25\% to 95\%.

Supplementary materials contain more tables and figures. We conducted a subjective study comparing the visual quality of purified images. The results showed that people prefer the quality of the three methods: Real-ESRGAN, DiffPure-based defences, and MPRNet. The supplementary materials also contain more detailed results of this study.

\section{Conclusion}

In this paper, we investigated the efficiency of adversarial purification methods as a defence from adversarial attacks on image quality assessment metrics. We conducted an extensive study involving 10 adversarial attacks and 16 purification methods and published a dataset of adversarial images \textit{link is hidden for a blind review}. 

The results showed that even simple and fast techniques like rotating or flipping the image can defeat attacks and maintain the visual quality of a purified image close to the original one. More complex defences can preserve and restore original quality: our proposed combinations of DiffPure with unsharp masking provide the highest SROCC of defended Linearity metric, keeping a high quality of purified images and a Gain score. The efficiency of tested purification defences behaves similarly for different iterative gradient-based attacks but differs for unrestricted attacks like AdvCf. Such attacks can be neutralized by training a neural network on adversarial examples. In our study, we proposed an FCN filter that efficiently defends Linearity from AdvCF attack.

\textbf{Limitations and future work.} In this study, we primarily focused on the defence of the Linearity metric. Despite the results being transferable to other NR IQA metrics as shown in the supplementary materials, there are more important issues than transferability that need to be studied in this field. For empirical attacks and defences, there is an eternal race: for each attack, we can propose a defence and so on. Developing provable defences for IQA metrics will be a promising direction for creating robust metrics, which is a subject of our future work.
 
%
%
\bibliographystyle{splncs04}
\bibliography{submission}
\end{document}